%% file: main.tex
\documentclass[10pt,twocolumn,letterpaper]{article}

\usepackage{wacv}              % To produce the CAMERA-READY version
\usepackage{multirow}
\usepackage[normalem]{ulem}
\usepackage{array}      % needed for \newcolumntype / \arraybackslash
\usepackage{multirow}   % needed for \multirow
\usepackage{pifont}
\newcommand{\xmark}{\ding{55}}  % ✗
\useunder{\uline}{\ul}{}

\definecolor{wacvblue}{rgb}{0.21,0.49,0.74}
\usepackage[pagebackref,breaklinks,colorlinks,allcolors=wacvblue]{hyperref}

\def\wacvPaperID{611} % *** Enter the WACV Paper ID here
\def\confName{WACV}
\def\confYear{2027}

\title{Learning Where and When: Patch-Based Spatiotemporal Localization in Weakly Supervised Video Anomaly Detection}

\author{Hamza Karim\\
University of South Florida\\
{\tt\small hamzakarim@usf.edu}
\and
Nghia Nguyen\\
University of South Florida\\
{\tt\small nghianguyen@usf.edu}
\and
Lokman Bekit\\
University of South Florida\\
{\tt\small Ibekit@usf.edu}
\and
Yasin Yilmaz\\
University of South Florida\\
{\tt\small yasiny@usf.edu}
}

\begin{document}
\maketitle
\input{sec/0_abstract}    
\input{sec/1_intro}
\input{sec/2_related_work}
\input{sec/3_methodology}
\input{sec/4_implement_details}
\input{sec/5_results}
\input{sec/6_ablation}
\input{sec/7_conclusion}
{
    \small
    \bibliographystyle{ieeenat_fullname}
    \bibliography{main}
}

\end{document}

%% file: sec/0_abstract.tex
\begin{abstract}
Weakly supervised video anomaly detection (WSVAD) has predominantly focused on temporal localization, identifying when anomalies occur while largely neglecting their spatial extent within frames. Yet, spatial localization is essential for interpretability and practical deployment in real-world settings. We introduce a patch-based spatiotemporal framework for weakly supervised anomaly localization that jointly models where and when anomalies occur. Our approach operates on grid-level patch features and learns region-level anomaly scores under a multiple instance learning paradigm. We further propose a Proximity-Aware Top-k spatiotemporal selection strategy that enables the model to generate fine-grained spatial anomaly maps without requiring bounding-box supervision during training. Our method surpasses existing state-of-the-art approaches across multiple benchmarks, yielding substantial gains in spatiotemporal localization accuracy. In addition, we release frame-level bounding-box annotations for the test sets of two widely used datasets, along with our code and pretrained models, providing new resources to facilitate future research in spatially grounded WSVAD.
\end{abstract}

%% file: sec/1_intro.tex
\section{Introduction}
\label{sec:intro}

Video anomaly detection aims to automatically identify events that deviate from expected patterns in long, untrimmed videos. It plays a central role in applications such as surveillance, public safety, traffic monitoring, and industrial inspection, where rapid and reliable identification of abnormal events is essential. While early approaches primarily focused on frame-level, coarse temporal localization, real-world deployments increasingly require spatiotemporal understanding, i.e., determining not only when an anomaly occurs but also where it manifests within each frame.

Obtaining fine-grained spatial annotations for anomalous events is expensive, time-consuming, and often subjective, especially in large-scale datasets. As a result, weakly supervised video anomaly detection (WSVAD), where models are trained using only coarse labels such as video-level annotations, has gained significant attention. The key challenge in this setting is to learn precise spatial localization signals despite the absence of explicit spatial or temporal supervision. Most existing weakly supervised methods primarily focus on temporal detection and rely on global representations, which limits their ability to capture localized abnormal patterns \cite{sultani2018real}, \cite{karim2024real},\cite{wu2024vadclip}, \cite{tian2021weakly}, \cite{joo2023clip}.

Detecting anomaly regions within each frame is crucial for several reasons. First, spatial localization greatly improves interpretability, enabling human operators to quickly verify alerts and understand the underlying cause of an anomaly. Second, it supports fine-grained decision making, such as tracking suspicious objects, triggering region-specific responses, or prioritizing computational resources in downstream systems. Third, spatial modeling provides richer contextual cues that help distinguish true anomalies from background clutter, reducing false positives that arise from back-ground biases.

In this work, we advance weakly supervised spatiotemporal video anomaly detection through the following contributions:
\begin{itemize}
    \item We propose a novel, patch-based weakly supervised spatiotemporal architecture that produces anomaly scores for \textbf{patches within each frame}, enabling fine-grained spatial localization without requiring bounding box supervision during training.

    \item We introduce a new \textbf{neighbor-aware top-$k$ MIL loss}, where the selected instances are formed through clustering based on temporal and spatial proximity, rather than conventional top-$k$ selection purely based on highest scores.

    \item We provide the \textbf{bounding box annotations} for the widely used anomaly detection datasets \textit{XD-Violence} and \textit{UBnormal} ~\cite{wu2020not, acsintoae2022ubnormal}, enabling standardized evaluation of spatial localization performance in the weakly supervised setting.
\end{itemize}

%% file: sec/2_related_work.tex
\section{Related Work}
\label{sec:related_work}
\subsection{Weakly Supervised Video Anomaly Detection}

Multiple instance learning (MIL) has become the dominant paradigm for weakly supervised video anomaly detection, where only video-level labels are available during training. The seminal work of Deep-MIL~\cite{sultani2018real} formulates anomaly detection as a ranking problem over temporal snippets, training a deep network to assign higher scores to anomalous bags than to normal ones. This framework established the core assumption that anomalous snippets exist within positive bags, and normal snippets dominate negative bags, and has since been extended in numerous directions.

Subsequent work addresses key limitations of the original ranking formulation. RTFM~\cite{tian2021weakly} improves discrimination by learning feature magnitudes that better separate anomalous instances, while incorporating temporal context through dilated convolutions and attention mechanisms.  URDMU~\cite{zhou2023dual} introduces dual memory units to model the distribution of normal and abnormal features more explicitly. %\textcolor{blue}
{DSANet \cite{yin2026dsanet} further addresses this by disentangling normal and abnormal features at coarse and fine-grained levels}. TCN-MIL~\cite{shao2023video} extends temporal modeling with non-local operations. CLIPTSA and VADCLIP ~\cite{joo2023clip, wu2024vadclip} leverage pretrained vision-language representations to align visual features with semantic descriptions, improving generalization to unseen categories. %\textcolor{blue}
{Recently, VERA ~\cite{Ye_2025_CVPR} adapts frozen VLMs for explainable anomaly detection without fine-tuning}.

Despite their strong temporal detection performance, these methods uniformly operate at the clip or snippet level, producing a single anomaly score per temporal unit. They offer no mechanism for identifying where within a frame an anomaly occurs, limiting their interpretability and practical utility in surveillance or safety-critical applications. Our work addresses this gap directly by reformulating MIL over spatial patches rather than temporal snippets, enabling joint spatiotemporal scoring under the same weak supervision.

\subsection{Spatiotemporal Anomaly Localization}

Spatial localization under weak supervision has received comparatively little attention, largely due to the absence of bounding-box annotations in standard benchmarks and the ambiguity of anomalous regions. Unsupervised reconstruction-based methods~\cite{hasan2016learning, zhao2017spatio} implicitly produce spatial error maps by measuring per-pixel or per-patch reconstruction residuals, but these signals are not grounded in semantic anomaly definitions and tend to be noisy in complex scenes. Attention-based unsupervised approaches~\cite{zhong2022bidirectional} identify candidate regions using multi-scale prediction error maps, but rely on heuristic selection rather than learned region-level discrimination.

A foundational step toward addressing the weakness in spatial localization was taken by Liu et al.~\cite{liu2019exploring}, who provided bounding-box annotations for the UCF-Crime dataset~\cite{sultani2018real} and introduced Temporal Intersection over Union (TIoU) as a standardized metric for evaluating spatiotemporal localization. Critically, their analysis also revealed that detectors frequently exploit background contextual cues rather than attending to true anomalous regions, a bias that persists in many subsequent methods and motivates architectures that explicitly model spatial evidence.

Building on this foundation, WSSTAD~\cite{wu2021weakly} proposes a dual-branch architecture for weakly supervised spatiotemporal tube localization, using only video-level labels. Through proposal reasoning and progressive refinement, it demonstrates that meaningful spatial grounding is achievable without bounding-box supervision during training. More recently, STPrompt~\cite{wu2024weakly} leverages vision-language prompt embeddings to align local spatial features with semantic anomaly descriptions, combining spatial and temporal streams to further reduce the background bias identified by~\cite{liu2019exploring}.

However, existing spatially aware methods share two fundamental limitations. First, they do not model fine-grained grid-level patch features under a MIL objective, which prevents the model from learning region-level anomaly scores in a principled, end-to-end manner. Second, their evaluation is limited due to the lack of spatial annotations in most benchmark datasets; currently, spatial annotations are available only for UCF-Crime and ShanghaiTech~\cite{zhang2016single}. To address this we propose a patch-based MIL architecture with a consecutive top-$k$ spatiotemporal selection strategy that produces explicit spatial anomaly maps. Furthermore, we provide new spatial annotations for XD-Violence and UBnormal, enabling a more comprehensive and standardized evaluation of spatial localization performance.

%% file: sec/3_methodology.tex
\section{Methodology}
\label{sec:methodology}

\begin{figure*}[h]
    \centering
    \includegraphics[width=\textwidth]{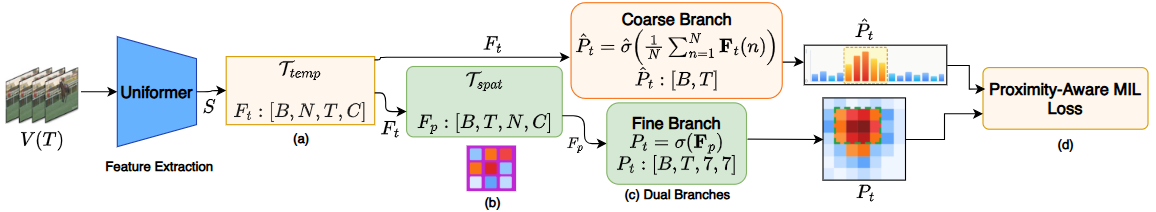}
    \caption{Overall architecture of the proposed method. The input video is processed through a feature extraction module, followed by (a) temporal and (b) spatial transformers to generate (c) coarse and fine-grained branch predictions, which are then evaluated using the (d) Proximity-Aware MIL Loss.}
    \label{fig:pipeline}
\end{figure*}

\subsection{Video Representation and Feature Extraction}

Given an input video $V$, we partition it into $T$ non-overlapping temporal segments, each containing 16 consecutive frames. The video tensor can therefore be represented as
\[
V \in \mathbb{R}^{T \times 16 \times H \times W \times 3}.
\]
Each segment is processed using a pretrained feature extractor (Uniformer\cite{li2023uniformer}) to obtain patch-level spatiotemporal representations
\[
\mathbf{S} \in \mathbb{R}^{B \times T \times H_p \times W_p \times C},
\]
where $B$ denotes the batch size, $C$ is the feature dimension, and $H_p = W_p = 7$. Consequently, the number of patches per segment is $N = H_p W_p = 49$.

During training, each mini-batch is constructed such that half of the samples correspond to anomalous videos and the remaining half corresponds to normal videos.

\subsection{Spatiotemporal Patch Aggregation}

We first aggregate the features $\mathbf{S}$ along the temporal dimension in order to model temporal dependencies for each spatial patch. Specifically, the features are reshaped into patch-wise temporal sequences
\[
\mathbf{X} \in \mathbb{R}^{B \times N \times T \times C}.
\]
Each patch sequence is processed by a temporal Transformer encoder
\[
\mathbf{F}_t = \mathcal{T}_{\text{temp}}(\mathbf{X}),
\]
where $\mathbf{F}_t \in \mathbb{R}^{B \times N \times T \times C}$
represents temporally aggregated patch features. The tensor is then rearranged to
$\mathbf{F}'_t \in \mathbb{R}^{B \times T \times N \times C}$,
and passed through a spatial Transformer encoder $\mathcal{T}_{\text{spat}}$ to capture spatial relationships among patches within each segment, yielding
\[
\mathbf{F}_p = \mathcal{T}_{\text{spat}}(\mathbf{F}'_t).
\]

\subsection{Coarse-Grained and Fine-Grained Branches}

The spatiotemporally aggregated features are subsequently used to construct two complementary branches: a coarse-grained (segment-level) branch and a fine-grained (patch-level) branch.

For the coarse-grained branch, the temporally aggregated features are first pooled across the spatial dimension to obtain a global representation for each segment:
\[
\mathbf{F}^g_t = \frac{1}{N} \sum_{n=1}^{N} \mathbf{F}_t(n) \in \mathbb{R}^{B \times T \times C}.
\]
This representation is then passed through a feed-forward network $\hat{\sigma}$ to produce segment-level anomaly scores
\[
\hat{P}_t = \hat{\sigma}(\mathbf{F}^g_t) \in \mathbb{R}^{B \times T}.
\]

As illustrated in Fig.~\ref{fig:pipeline}, in parallel, the fine-grained branch applies another feed-forward network $\sigma$ to the output of spatial Transformer to generate patch-level anomaly probabilities:
\[
P_t = \sigma(\mathbf{F}_p) \in \mathbb{R}^{B \times T \times 7 \times 7}.
\]
\subsection{Proximity-Aware MIL Loss}

We use the coarse and fine-grained predictions $\hat{P}_t$ and $P_t$ as inputs to a novel multiple instance learning (MIL) loss. 
Unlike conventional MIL formulations that select the top-$K$ instances solely based on their individual anomaly scores,  we introduce a proximity-aware selection strategy motivated by the observation that anomalous events typically occur over temporally contiguous intervals. %\textcolor{blue}
{Note that this proximity prior is imposed only during training; at inference the model scores each patch and segment independently, so spatially or temporally distinct anomalies can still be localized separately. Closely proximate anomalous regions may, however, merge into a single response, reflecting a deliberate trade-off between spatial smoothness and separation.}

\begin{figure*}[h]
    \centering
    \includegraphics[width=.8\linewidth]{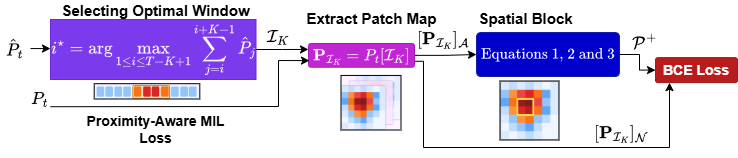}
    \caption{Detailed illustration of the Proximity-Aware MIL Loss module, showing the optimal window selection, patch map extraction, and spatial block computation leading to the final BCE Loss.}
    \label{fig:mil_loss}
\end{figure*} 

Considering $B=1$, given the segment-level anomaly scores 
\[
\hat{P_t} = [\hat{P}_1, \ldots, \hat{P}_T] \in \mathbb{R}^{T},
\]
we perform sliding-window selection with window length $K$ and stride $1$. 
For each candidate window starting at time index $i$, we compute the cumulative score
\[
s_i = \sum_{j=i}^{i+K-1} \hat{P}_j, 
\quad i = 1, \ldots, T-K+1.
\]
The optimal window is obtained by
\[
i^\star = \arg\max_{i} \; s_i.
\]

The selected top-$K$ segment indices and their corresponding scores are then defined as
\[
\mathcal{I}_K = \{\, i^\star, i^\star+1, \ldots, i^\star+K-1 \,\},
\quad
\hat{P}_{\mathcal{I}_K} = \{\, \hat{P}_t \mid t \in \mathcal{I}_K \,\}.
\]
We apply standard binary cross-entropy (BCE) loss to $\hat{P}_{\mathcal{I}_K}$ based on the video label:
\begin{align}
\mathcal{L}_{\text{temp}}^{\text{anom}} &= \frac{1}{|\hat{P}_{\mathcal{I}_K}|}
\sum_{p \in \hat{P}_{\mathcal{I}_K}}
\mathrm{BCE}(p, 1), \nonumber\\
\mathcal{L}_{\text{temp}}^{\text{norm}} &= \frac{1}{|\hat{P}_{\mathcal{I}_K}|}
\sum_{p \in \hat{P}_{\mathcal{I}_K}}
\mathrm{BCE}(p, 0). \nonumber
\end{align}
To maintain temporal–spatial consistency, we reuse the selected segment index set $\mathcal{I}_K$ 
to extract the corresponding patch-level predictions from $P_t$:
\[
%\mathbf{P}_{\mathcal{I}_K} = P_t[\mathcal{I}_K] \in \mathbb{R}^{K \times 7 \times 7}.
P_{\mathcal{I}_K} = \{P_t | t\in \mathcal{I}_K \} \in \mathbb{R}^{K \times 7 \times 7}.
\]
%The selected patches constitute the positive instance bag for the fine-grained MIL objective.
%For videos belonging to this anomalous bag, 
For the fine-grained MIL objective, 
we further identify the most discriminative spatial region within each selected segment as can be seen in Fig ~\ref{fig:mil_loss}. 
Specifically, for each $k \in \{1,\dots,K\}$, we slide a convolutional window of spatial size $z \times z$ over the $7 \times 7$ patch grid and compute the summed response:
\begin{equation}
r^{(k)}_{u,v} = 
\sum_{i=u}^{u+z-1} \sum_{j=v}^{v+z-1}
P_{\mathcal{I}_K}(k,i,j),
\end{equation}
for all valid spatial locations $(u,v)$. The most anomalous spatial region is selected as
\begin{equation}
(u^\star_k, v^\star_k) = 
\arg\max_{u,v} \; r^{(k)}_{u,v}.
\end{equation}
The corresponding $Kz^2$ patch scores form the positive instance set
%\textcolor{blue}
{\begin{equation}
\mathcal{P}^{+} =
\bigcup_{k=1}^{K}
\bigl\{\,
P_{\mathcal{I}_K}(k,i,j)
\;\big|\;
(i,j)\in \mathcal{W}_z(u^\star_k, v^\star_k)
\,\bigr\},
\end{equation}
where $\mathcal{W}_z(u,v) = [u, u{+}z{-}1]\times[v, v{+}z{-}1]$ 
denotes the $z\times z$ spatial block anchored at $(u,v)$}. For anomalous videos, we apply the BCE loss against the positive label:
\[
\mathcal{L}_{\text{patch}}^{\text{anom}}
=
\frac{1}{|\mathcal{P}^{+}|}
\sum_{p \in \mathcal{P}^{+}}
\mathrm{BCE}(p, 1).
\]

For normal videos, all patches are treated as negative instances and supervised with label $0$:
\[
\mathcal{L}_{\text{patch}}^{\text{norm}}
=
\frac{1}{49K}
\sum_{k=1}^{K}
\sum_{i=1}^{7}
\sum_{j=1}^{7}
\mathrm{BCE}\!\left(
P_{\mathcal{I}_K}(k,i,j),\, 0
\right).
\]
To encourage compact anomaly regions and suppress noise, we impose %sparsity loss on $\mathbf{P}_{\mathcal{I}_K}$ 
constraints
for both anomalous and normal sets. For an anomalous set, we use L1-norm to introduce sparsity in the predicted patches while we use L2-norm to ensure consistency between all scores within the normal set since $P_t$ in the normal set should be close to 0 for all patches:

\begin{equation}
\mathcal{L}_{\text{sparse}} =
\frac{1}{|\mathcal{A}|}\sum_{b \in \mathcal{A}}
\left\|
{P}_{\mathcal{I}_K}^{(b)}
\right\|_1
+
\frac{1}{|\mathcal{N}|}\sum_{b \in \mathcal{N}}
\left\|
{P}_{\mathcal{I}_K}^{(b)}
\right\|_2,
\end{equation}
where $\mathcal{A}$ and $\mathcal{N}$ denote anomalous and normal sets.
The final training loss is
\begin{equation}
\mathcal{L} =
\lambda_t (\mathcal{L}_{\text{temp}}^{\text{anom}} + \mathcal{L}_{\text{temp}}^{\text{norm}})
+ \lambda_s
(\mathcal{L}_{\text{spat}}^{\text{anom}} + \mathcal{L}_{\text{spat}}^{\text{norm}})
+ \lambda_{sp} \mathcal{L}_{\text{sparse}},
\end{equation}
where the constants $\lambda_t$, $\lambda_s$, and $\lambda_{sp}$ control the weights of temporal, spatial, and sparse losses. 

%% file: sec/4_implement_details.tex
\section{Implementation Details and Evaluation Metrics}
\label{sec:implement_details}

We evaluate our approach against state-of-the-art methods using two complementary metrics: temporal Intersection over Union (TIoU) and Track-Based Detection Rate / Region-Based Detection Rate (TBDR/RBDR). While prior weakly supervised video anomaly detection works predominantly report TIoU, this metric mainly reflects spatial alignment and does not fully capture the spatiotemporal localization capability of models, e.g., false positives. To provide a more comprehensive evaluation, we additionally report TBDR/RBDR, which jointly assesses detection performance across both spatial and temporal dimensions. For training, we use the Adam optimizer with an initial learning rate of $0.001$. The loss balancing coefficients are set to $\lambda_t = 1$, $\lambda_s = 0.1$, and $\lambda_{sp} = 0.01$ and spatial window size $z=2$. %\textcolor{blue}
{In line with recent literature \cite{pu2024learning},\cite{wu2024vadclip}, we set $K=(T/16)+1$.}

\subsection{Baseline Implementations}
%\textcolor{blue}
{Spatiotemporal localization under weak supervision remains underexplored. There are only a few methods target this exact setting, which limits the pool of directly comparable approaches. To ensure a fair and comprehensive evaluation, we therefore implement several competitive baselines on our UniFormer patch features. %Our reported gains stem from two sources: a stronger backbone, patch-level UniFormer features versus the clip-level I3D/C3D features used by prior work, and the localization strategy itself, as illustrated by ST-Prompt, which produces a single bounding box from spatial heatmaps where our method performs explicit patch-level localization. 
A fully fair comparison with published methods is further complicated by the absence of public code or pretrained models; this also accounts for the N/A entries in Table 1, since prior works were evaluated only on UCF-Crime and ShanghaiTech and cannot be reliably extended to XD-Violence or UBNormal. To support reproducibility, we release our code, pretrained models, extracted features, and ground-truth labels for %and we evaluate on four widely used WSVAD benchmarks: \textit{UCF-Crime}, \textit{ShanghaiTech}, 
the \textit{UBnormal}, and \textit{XD-Violence} datasets.}

\paragraph{Deep-MIL.}
A simple patch-based MIL variant in which patch-level features are independently processed using a feed-forward network, and optimization is performed under a standard MIL objective. This baseline evaluates the impact of patch-level reasoning without explicit spatiotemporal interaction modeling.

\paragraph{TCN-MIL.}
Inspired by temporal convolutional modeling, we implement a TCN-based MIL approach~\cite{shao2023video} in which 1D temporal convolutions are applied across segment sequences to aggregate temporal dependencies for each patch. This allows the model to capture local temporal context while maintaining computational efficiency.

\paragraph{LSTM-MIL.}
To model sequential dependencies explicitly, we implement an LSTM-based MIL baseline inspired from ~\cite{li2020multi}, where patch features are processed through an LSTM module along the temporal dimension. This enables recurrent modeling of temporal evolution within each spatial location.

\paragraph{GCN-MIL.}
Finally, we implement a graph-based MIL baseline~\cite{zhong2019graph} that combines temporal convolution for modeling temporal dynamics with a Graph Convolutional Network (GCN) to capture spatial relationships among patches within each segment. This baseline explicitly models both temporal and spatial dependencies through structured relational learning.

\subsection{Evaluation Metrics}
We evaluate our method using two metrics: temporal Intersection-over-Union (TIoU) and the Track-Based / Region-Based Detection Rate (TBDR/RBDR) criteria.

The \textbf{TIoU} metric measures the overlap between predicted and ground-truth anomaly regions within each frame containing an anomaly, providing a standard evaluation of spatial localization accuracy. While it is the only metric used in the literature \cite{liu2019exploring}, \cite{wu2021weakly}, \cite{wu2024weakly}, TIoU evaluates only \emph{where} an anomaly occurs with in a frame, and does not account for the temporal inaccuracy of the model. As a result, a method may achieve high TIoU by correctly identifying anomalous spatial regions in ground-truth positive frames while ignoring false detections in frames with no anomalies, i.e., by raising abundance of anomaly alarms most of which are false. Therefore, TIoU alone is insufficient for evaluating spatiotemporal anomaly localization methods that aim to precisely localize abnormal regions.

To address this limitation, we additionally adopt the TBDR/RBDR evaluation protocol introduced in~\cite{ramachandra2020street}. These criteria are specifically designed for region-level and track-level anomaly detection and provide a more comprehensive assessment of spatial and temporal localization quality by considering false detections as well. 

The \textbf{Region-Based Detection Rate (RBDR)} measures the proportion of ground-truth anomalous regions that are successfully detected with an IoU exceeding a predefined threshold ($\alpha$) while also checking number of false positives (with and IoU less than the predefined threshold) per frame. 
This metric evaluates spatial localization performance at the frame level.

The \textbf{Track-Based Detection Rate (TBDR)} extends this evaluation temporally by considering complete anomalous tracks (i.e., sequences of spatially localized anomaly regions across time). A track is considered correctly detected if a sufficient fraction ($\beta$) of its constituent regions are matched by predictions above the IoU threshold. Consequently, TBDR captures spatiotemporal consistency and penalizes fragmented or sporadic detections.

By incorporating both TIoU and TBDR/RBDR, our evaluation protocol assesses not only the temporal extent of anomalies but also the precision and consistency of their spatial localization, leading to a more rigorous and application-relevant evaluation of spatiotemporal anomaly detection performance. For this work we set values of $\alpha$ and $\beta$ to 0.1, as in~\cite{ramachandra2020street}.

%% file: sec/5_results.tex
\section{Results}
\label{sec:results}

\begin{figure*}[t]
  \centering
  \includegraphics[width=\textwidth]{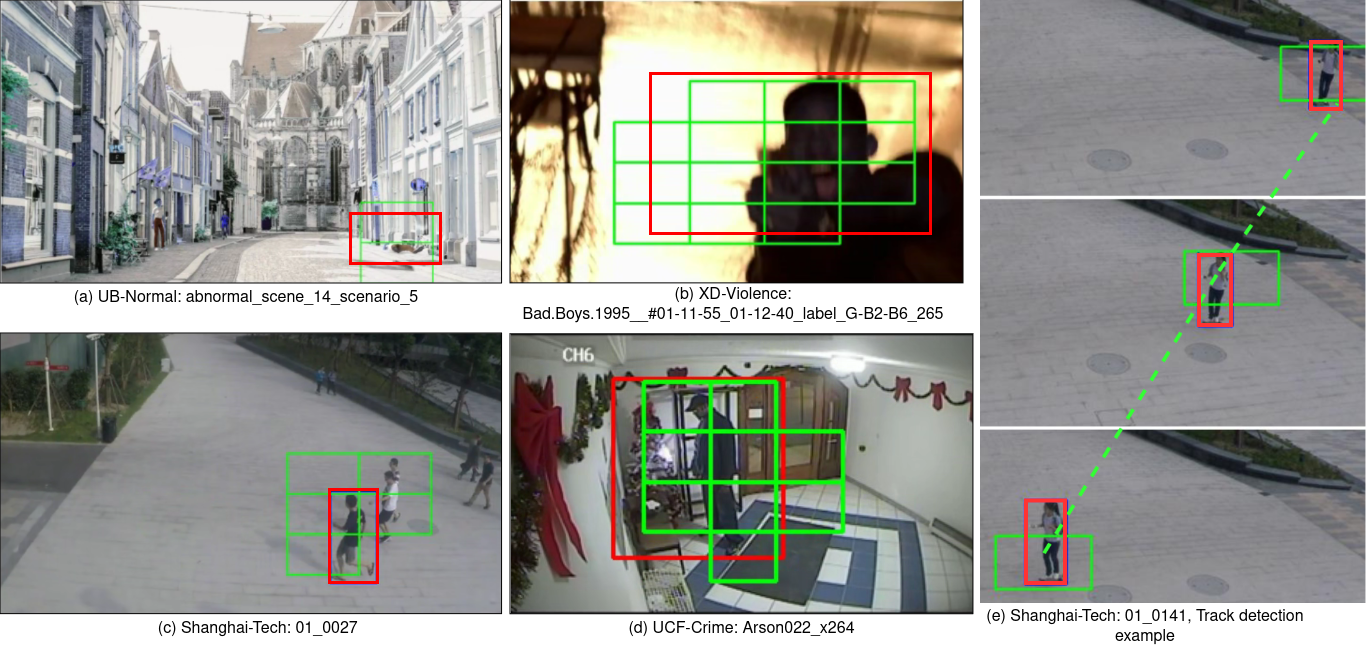}
  \caption{(a)-(d) Examples of detected patches (green boxes) and ground truth
  (red box) from the four datasets. (e) Example of detected track across multiple frames.}
  \label{fig:example}
\end{figure*}

In addition to the quantitative performance evaluation in the following sections, we also present some spatial anomaly localization samples of our method in Fig. \ref{fig:example} for qualitative assessment. 

\subsection{TIoU Performance}
Table~\ref{tab:TIoU} reports the performance in terms of TIoU across four benchmark datasets. Our method consistently outperforms all prior approaches by a significant margin. On UCF-Crime, our model, surpasses the strongest baseline (LSTM-MIL) by nearly 7.6\%. A similar trend is observed on ShanghaiTech, improving over ST-Prompt by 2.79\%.\par  
On XD-Violence, our approach reaches \textbf{39.58}\%, outperforming LSTM-MIL and TCN-MIL, demonstrating robust temporal boundary estimation even in highly diverse anomaly scenarios. Similarly, improvement observed on UB-Normal is also significant at a 6.25\% difference.

\begin{table}[t]
\centering
\caption{TIoU comparison against some state-of-the-art models on four popular datasets.}
\label{tab:TIoU}
\resizebox{\columnwidth}{!}{%
\begin{tabular}{|l|c|c|c|c|}
\hline
 & UCF-Crime & ShanghaiTech & XD-Violence & UB-Normal \\ \hline
Deep-MIL  & 16.82 & 2.46 & 32.69 & 9.44 \\ \hline
Liu et al.& 16.40 & N/A  & N/A   & N/A  \\ \hline
WSSTAD    & 8.98  & N/A  & N/A   & N/A  \\ \hline
VADCLIP   & 22.05 & 4.09 & N/A   & N/A  \\ \hline
ST-Prompt & 23.90 & \underline{9.77} & N/A & N/A \\ \hline
GCN-MIL   & 25.56 & 5.63 & 32.78 & 9.44 \\ \hline
TCN-MIL   & 27.98 & 7.55 & 33.66 & 10.68 \\ \hline
LSTM-MIL  & \underline{32.91} & 4.85 & \underline{37.79} & \underline{13.51} \\ \hline
\textbf{OURS} & \textbf{40.51} & \textbf{12.56} & \textbf{39.58} & \textbf{19.76} \\ \hline
\end{tabular}%
}
\end{table}

\subsection{Region and Track-Based Evaluation (TBDR/RBDR)}
To further evaluate spatial localization capability, Table~\ref{tab:TBDR} presents results using TBDR and RBDR metrics. Our approach achieves the best TBDR across all four datasets, confirming its superior ability to correctly detect anomalous tracks over time.

\begin{table*}[t]
\centering
\caption{TBDR/RBDR comparison against some state-of-the-art models on four popular datasets.}
\label{tab:TBDR}
\begin{tabular}{|c|cc|cc|cc|cc|}
\hline
\multirow{2}{*}{Model} & \multicolumn{2}{c|}{UCF-Crime}                       & \multicolumn{2}{c|}{XD-Violence}                     & \multicolumn{2}{c|}{UB-Normal}                       & \multicolumn{2}{c|}{hanghai Tech}                    \\ \cline{2-9} 
                       & \multicolumn{1}{c|}{TBDR}           & RBDR           & \multicolumn{1}{c|}{TBDR}           & RBDR           & \multicolumn{1}{c|}{TBDR}           & RBDR           & \multicolumn{1}{c|}{TBDR}           & RBDR           \\ \hline
Deep-MIL               & \multicolumn{1}{c|}{67.01}          & 62.52          & \multicolumn{1}{c|}{67.26}          & 49.35          & \multicolumn{1}{c|}{40.31}          & 19.67          & \multicolumn{1}{c|}{12.07}          & 5.33           \\ \hline
GCN-MIL                & \multicolumn{1}{c|}{69.71}          & 62.52          & \multicolumn{1}{c|}{73.71}          & 65.64          & \multicolumn{1}{c|}{45.86}          & 20.53          & \multicolumn{1}{c|}{14.56}          & 7.42           \\ \hline
TCN-MIL                & \multicolumn{1}{c|}{70.1}           & 66.83          & \multicolumn{1}{c|}{69.81}          & 62.52          & \multicolumn{1}{c|}{47.54}          & 24.47          & \multicolumn{1}{c|}{{\ul 16.98}}    & {\ul 9.04}     \\ \hline
LSTM-MIL               & \multicolumn{1}{c|}{{\ul 72.26}}    & 67.87          & \multicolumn{1}{c|}{{\ul 74.10}}    & \textbf{69.82} & \multicolumn{1}{c|}{{\ul 45.09}}    & {\ul 29.48}    & \multicolumn{1}{c|}{11.31}          & 5.16           \\ \hline
OURS                   & \multicolumn{1}{c|}{\textbf{77.20}} & \textbf{69.83} & \multicolumn{1}{c|}{\textbf{78.66}} & {\ul 67.02}    & \multicolumn{1}{c|}{\textbf{57.23}} & \textbf{33.29} & \multicolumn{1}{c|}{\textbf{33.30}} & \textbf{19.19} \\ \hline
\end{tabular}
\end{table*}

On UCF-Crime, our model attains a TBDR of \textbf{77.20}\% and an RBDR\% of \textbf{69.83}\%, outperforming LSTM-MIL by 4.94\% and 1.96\% points respectively. For XD-Violence, we achieve the highest TBDR of \textbf{78.66}\%, demonstrating strong robustness in complex scenes with multiple anomaly types, while maintaining competitive region accuracy.  

On ShanghaiTech, our method shows substantial gains, achieving \textbf{33.30}\% TBDR and \textbf{19.19}\% RBDR, which represents a large improvement over prior methods and highlights the effectiveness of our patch-level modeling for dense scenes. Another notable improvement appears on UB-Normal, where we obtain \textbf{57.23\%} TBDR and \textbf{33.29}\% RBDR, significantly outperforming all baselines and demonstrating strong generalization to subtle and small-scale anomalies.

\subsection{Model Computational Efficiency and Runtime Comparison}
We evaluate the computational efficiency of all models on a 60-frame input in Table \ref{tab:memory_runtime}, where the time and memory consumption include the shared UniFormer backbone (40 ms, 5 GB). Our model requires 44.23 ms and 6.747 GB, slightly higher than the baselines due to the additional spatial–temporal refinement modules, yet remains suitable for real-time inference.
\begin{table}[t]
\centering
\caption{Computational and Runtime Comparison of Models.}
\label{tab:memory_runtime}
\begin{tabular}{|c|c|c|}
\hline
Baseline & Time (ms)/60 fr. & Memory (GB) \\ \hline
LSTM-MIL       & 41.755  & 6.618 \\ \hline
GCN            & 41.133  & 6.584 \\ \hline
Sultani et al. & 41.62   & 6.572 \\ \hline
Our model      & 44.23   & 6.747 \\ \hline
\end{tabular}
\end{table}

\subsection{Zero-Shot Performance}

To evaluate the generalization ability of the proposed model, we conduct a zero-shot anomaly detection analysis across different anomaly labels. Specifically, for each selected anomaly class in the UCF-Crime dataset, the model is trained without any samples from that class, and then evaluated exclusively on the unseen class during testing. This setup simulates a realistic deployment scenario suitable with the nature of anomaly detection problem in which certain types of anomaly may not be observed during training but must still be detected at inference time.

Table \ref{tab:TBDRDRop} reports the TBDR/RBDR performance drop between the fully trained model (trained on all anomaly categories) and the zero-shot setting where the target class is unseen during training. The results are averaged over the four anomaly labels with the largest number of frames in the test set (Explosion, Shooting, Shoplifting, and RoadAccidents). We then compute the average percentage decrease in TBDR and RBDR to quantify the impact of zero-shot generalization. A larger performance drop indicates that the model relies more heavily on class-specific patterns learned during training, whereas a smaller drop suggests stronger generalization to previously unseen anomaly categories.

\begin{table}[h]
\centering
\caption{Comparison of baseline models zero-shot performance on UCF-Crime}
\begin{tabular}{|c|c|}
\hline
Model                  & TBDR/RBDR Drop (\%)  \\ \hline
Deep-MIL      & \underline{5.02}/\underline{3.50}     \\ \hline
TCN-MIL          & 9.12/7.55       \\ \hline
 GCN    & 5.80/\textbf{3.00}      \\ \hline
 LSTM      & 12.94/8.3     \\ \hline
\multicolumn{1}{|l|}{\textbf{Our model}} & \textbf{3.08}/4.4  \\ \hline
\end{tabular}
\label{tab:TBDRDRop}
\end{table}
\subsection{Annotations for XD-Violence and UB-Normal test datasets}
\label{anno}
Following the annotation protocol of Liu et al. \cite{liu2019exploring}, we annotate every 16-th frame of each anomaly track in the test videos. For each selected frame, we manually draw bounding boxes that tightly enclose the anomalous regions. Using this protocol, we create spatial annotations for the test sets of two widely used video anomaly detection benchmarks: XD-Violence and UBNormal. For the XD-Violence dataset, we annotate 500 anomalous test videos, resulting in 27,654 frames with bounding-box labels. For the UBNormal test set, we annotate 158 anomalous videos, producing 3,388 annotated frames with spatial annotations.

All annotations are included in the supplementary material as annotation.zip. The archive contains bounding-box annotation files organized by video ID and frame index. A README file is also provided to describe the annotation format and to give instructions for loading and using the annotations during evaluation.

%% file: sec/6_ablation.tex
\section{Ablation Study}
\label{sec:ablation}

\subsection{Component Ablation}

Our patch-based model is a lightweight spatiotemporal transformer architecture that operates on patch-level video features to produce anomaly scores. It consists of three main components: a spatial transformer for modeling intra-frame relationships, a temporal transformer for capturing temporal dynamics, and a temporal head for motion-aware anomaly scoring.

In this ablation study, we systematically remove or modify these components, along with the Proximity MIL mechanism, to evaluate their individual contributions. In all ablation tests, performance is measured using TBDR and RBDR to quantify the impact of each component on spatio-temporal anomaly localization. %TBDR measures whether a certain percentage of frames in an anomaly track is correctly localized, while RBDR measures per-frame region overlap; thus, RBDR is a stricter, more spatially-focused metric.%
\begin{table}[h]
\centering
\footnotesize
\setlength{\tabcolsep}{4pt}
\caption{Impact of Model Components on TBDR/RBDR Performance.}
\label{tab:component_ablation}
\begin{tabular}{|c|c|c|c|c|}
\hline
$\mathcal{T}_{\text{temp}}$ & $\mathcal{T}_{\text{spat}}$ & \shortstack{Proximity\\MIL} & \shortstack{TBDR/RBDR\\(\%)} & \shortstack{TBDR/RBDR\\Drop (\%)}\\ \hline
\xmark     & \checkmark & \checkmark & 71.19/69.00 & 6.01/0.83 \\ \hline
\checkmark & \xmark     & \checkmark & 58.05/47.00 & 19.15/22.83 \\ \hline
\checkmark & \checkmark & \xmark     & 72.86/68.08 & 4.34/1.75 \\ \hline
\checkmark & \checkmark & \checkmark & \textbf{77.20/69.83} & -/- \\ \hline
\end{tabular}
\end{table}

The results in Table \ref{tab:component_ablation} demonstrate that each component contributes significantly to the overall performance. Removing the spatial transformer results in the largest performance decrease in TBDR (19.15) and a substantial drop in RBDR (22.83), showing that spatial attention is crucial for precise region-level localization, while also contributing to anomaly detection performance. In contrast, removing the temporal transformer leads to a significant decrease in TBDR (6.01), highlighting the critical role of temporal modeling in capturing anomalies . The removal of the proximity-based MIL results in comparatively smaller drops (4.34 in TBDR and 1.75 in RBDR), yet still important in refining spatial–temporal alignment under weak supervision. Overall, the complete model achieves the best performance in both TBDR and RBDR, confirming that the proposed components are complementary and jointly contribute to robust anomaly localization.

\subsection{Ablation Study Of Proximity MIL On State-of-the-art Models}
Next, we evaluate the effectiveness and generalizability of the proposed Proximity MIL loss by replacing the standard MIL objective in several representative weakly supervised video anomaly detection frameworks, including URDMU, PEL4VAD, VADCLIP, and the baseline model of Deep-MIL. For each method, we keep the original model architecture, and modifying only the MIL loss formulation to ensure a fair comparison. All experiments are conducted on the UCF-Crime dataset, independently trained ten times with different random seeds. We then report the mean AUC and standard deviation across runs.

\begin{table}[h]
\centering
\footnotesize
\setlength{\tabcolsep}{5pt}
\renewcommand{\arraystretch}{1.1}
\caption{Standard vs. Proximity MIL on UCF-Crime.}
\label{tab:baseline_ablation}
\begin{tabular}{lccc}
\hline
\multirow{2}{*}{Model} & Standard MIL & Proximity MIL & \multirow{2}{*}{$\Delta$AUC} \\ \cline{2-3}
 & Mean AUC (SD) & Mean AUC (SD) & \\ \hline
Deep-MIL~\cite{sultani2018real} & 77.7 (0.45) & \textbf{78.2} (0.46) & +0.5 \\
URDMU~\cite{zhou2023dual} & 85.52 (0.744) & \textbf{86.02} (0.508) & +0.5 \\
PEL4VAD~\cite{pu2024learning} & 85.21 (0.366) & \textbf{85.35} (0.435) & +0.14 \\
VADCLIP~\cite{wu2024vadclip} & 86.26 (0.918) & \textbf{86.41} (0.612) & +0.15 \\ \hline
\end{tabular}
\end{table}

As shown in Table \ref{tab:baseline_ablation}, the proposed Proximity MIL loss consistently improves performance across all evaluated models, yielding AUC gains ranging from 0.14\% to 0.50\%. Consistent improvements across both classical MIL-based frameworks (e.g., Deep-MIL) and modern transformer-based approaches (e.g., PEL4VAD and VADCLIP) demonstrate the strong generalization ability of the proposed loss function. By encouraging smoother temporal anomaly scores and mitigating noisy top-k instance selection, the Proximity MIL provides a more reliable optimization target compared to standard MIL, which helps improve robustness and model performance. 

\subsection{Ablation Study on Spatial Window Size, Patch Grid Resolution, and Normalization Losses}

\textbf{}Table~\ref{tab:grid_window_ablation} reports the effect of patch grid resolution and spatial window size $z$. The $7{\times}7$ grid with $z=2$ achieves the best performance (TBDR: 77.20\%, RBDR: 69.83\%), as coarser grids lose spatial detail while finer grids introduce noise; similarly, a moderate window size captures sufficient context without over-smoothing. Thus, this configuration is adopted as the default setting.

Table~\ref{tab:loss_ablation} ablates the normalization losses. Using both $\mathcal{L}_1$ and $\mathcal{L}_2$ together yields the best results, while each loss alone impacts negatively, confirming their complementarity.

\begin{table}[t]
\footnotesize
\centering
\renewcommand{\arraystretch}{1.3}
\setlength{\tabcolsep}{4pt}

\caption{Grid and window size ablation (TBDR/RBDR \%).}
\label{tab:grid_window_ablation}
\begin{tabular}{|c|c|c|c|}
\hline
\multirow{2}{*}{\textbf{Grid}} & \multicolumn{3}{c|}{\textbf{Window size}} \\
\cline{2-4}
 & $z=1$ & $z=2$ & $z=3$ \\
\hline
$6\times6$ & 70.26/65.62 & 76.68/68.45 & 73.15/67.86 \\
\hline
$7\times7$ & 72.41/63.34 & \textbf{77.20}/\textbf{69.83} & 76.46/68.44 \\
\hline
$8\times8$ & 67.06/65.48 & 75.38/67.84 & 75.42/68.85 \\
\hline
\end{tabular}

\vspace{1em}

\caption{Normalization loss ablation.}
\label{tab:loss_ablation}
\begin{tabular}{|c|c|}
\hline
\textbf{Configuration} & \textbf{TBDR/RBDR} \\
\hline
Base & 75.79/68.96 \\
\hline
$\mathcal{L}_1$ only & 73.33/67.50 \\
\hline
$\mathcal{L}_2$ only & 74.67/68.46 \\
\hline
$\mathcal{L}_1+\mathcal{L}_2$ & \textbf{77.20}/\textbf{69.83} \\
\hline
\end{tabular}
\end{table}

%% file: sec/7_conclusion.tex
\section{Conclusion}
In this work, we addressed a fundamental limitation of weakly supervised video anomaly detection by moving beyond purely temporal localization toward joint spatiotemporal anomaly detection. We proposed a patch-based spatiotemporal framework that reformulates MIL at the spatial patch level, enabling region-level anomaly scoring without requiring bounding-box supervision during training. Our proximity-aware top-K selection encourages the model to focus on anomaly segments that occur close together in time and space, producing more stable and interpretable localization results.

Extensive experiments on UCF-Crime, ShanghaiTech, XD-Violence, and UBnormal demonstrate consistent improvements in both TIoU and TBDR/RBDR metrics. Ablation studies confirm the complementary roles of temporal modeling, spatial attention, and the proposed Proximity MIL loss. Furthermore, the framework introduces only modest computational overhead, maintaining practical efficiency.

Finally, by releasing bounding-box annotations for XD-Violence and UBnormal, we aim to facilitate standardized evaluation and encourage further research in weakly supervised anomaly detection.